\documentclass[9pt]{article}

\usepackage{geometry}
\usepackage{microtype}
\usepackage{subfigure}
\usepackage{booktabs} 

\usepackage{multirow}
\usepackage{graphicx}
\usepackage{amsmath,amsthm}
\theoremstyle{definition}

\usepackage{xspace,color}
\usepackage{diagbox,multirow} 
\usepackage{subfigure}
\usepackage{tikz,bm}
\usepackage{algorithmic,algorithm}
\definecolor{PineGreen}{rgb}{0.0, 0.47, 0.44}
\usepackage{enumitem}
\setlist[itemize]{leftmargin=*}
\usepackage{amsfonts}
\usepackage[utf8]{inputenc}
\usepackage[T1]{fontenc}    
\usepackage{url}            
\usepackage{booktabs}       
\usepackage{nicefrac}       
\usepackage{microtype}      
\usepackage{tikz,bm}
\usepackage{adjustbox}      
\usepackage{filecontents}
\usepackage{array,rotating}
\usepackage[labelfont=bf]{caption}

\usepackage{caption}
\usepackage{xspace,color}
\usepackage{enumerate}
\usepackage{enumitem,amsmath,amssymb}
\usepackage{hyperref}

\usepackage{amssymb,bbm}
\usepackage{amsmath,amsthm,amsfonts,booktabs}
\usepackage[switch]{lineno}

\theoremstyle{definition}

\begin{document}
%
\title{\texttt{TWIN-GPT}: Digital Twins for Clinical Trials via Large Language Model}

\author{Yue Wang$^{1*}$ \and Tianfan Fu$^{2*}$ \and Yinlong Xu$^{1}$ \and Zihan Ma$^{1}$ \and Hongxia Xu$^{1}$ \and Bang Du$^{1}$ \and Yingzhou Lu$^3$ \and Honghao Gao$^{4}$ \and Jian Wu$^{1}$ \and Jintai Chen$^{5,+}$}
\date{%
    $^1$Second Affiliated Hospital School of Medicine, School of Public Health, Zhejiang University \\
    $^2$Computer Science Department, Rensselaer Polytechnic Institute \\
    $^3$School of Medicine, Stanford University \\
    $^4$School of Computer Engineering and Science, Shanghai University \\ 
    $^5$Computer Science Department, University of Illinois Urbana-Champaign \\ 
    $+$ Corresponding author \\ 
}
\maketitle

\begin{abstract}
Clinical trials are indispensable for medical research and the development of new treatments. However, clinical trials often involve thousands of participants and can span several years to complete, with a high probability of failure during the process. Recently, there has been a burgeoning interest in virtual clinical trials, which simulate real-world scenarios and hold the potential to significantly enhance patient safety, expedite development, reduce costs, and contribute to the broader scientific knowledge in healthcare. Existing research often focuses on leveraging electronic health records (EHRs) to support clinical trial outcome prediction. Yet, trained with limited clinical trial outcome data, existing approaches frequently struggle to perform accurate predictions. Some research has attempted to generate EHRs to augment model development but has fallen short in personalizing the generation for individual patient profiles. Recently, the emergence of large language models has illuminated new possibilities, as their embedded comprehensive clinical knowledge has proven beneficial in addressing medical issues. In this paper, we propose a large language model-based digital twin creation approach, called \texttt{TWIN-GPT}. \texttt{TWIN-GPT} can establish cross-dataset associations of medical information given limited data, generating unique personalized digital twins for different patients, thereby preserving individual patient characteristics. Comprehensive experiments show that using digital twins created by \texttt{TWIN-GPT} can boost the clinical trial outcome prediction, exceeding various previous prediction approaches. Besides, we also demonstrate that \texttt{TWIN-GPT} can generate high-fidelity trial data that closely approximates specific patients, aiding in more accurate result predictions in data-scarce situations. Moreover, our study provides practical evidence for the application of digital twins in healthcare, highlighting its potential significance.
\end{abstract}

\section{Introduction}


Clinical trials serve as essential research investigations in the field of prospective medicine. They are designed to rigorously evaluate the safety and efficacy of new treatments (e.g., drugs, or medical devices).
Despite playing a pivotal role in advancing medical knowledge and enhancing patient care, these trials typically engage tens to thousands of human participants and frequently extend over several years, hindering the progress of research and development.
There is an increasing interest in leveraging artificial intelligence models to predict trial outcomes, leveraging electronic health records (EHR) and significantly accelerating the pace of research~\cite{ren2019deep,choi2017using,fu2022hint,chen2024uncertainty}. 
Learning from EHR data, models are able to identify the relationship between potential health issues with personalized characteristics, thus excelling at forecasting disease occurrence and even developing personalized treatment plans~\cite{shen2023genocraft,lu2019integrated}.



Existing approaches leveraging EHR for clinical trial outcome prediction still face a set of challenges. These methods often fail to account for individual differences between patients and variations among diseases, resulting in predictions that may not align with real-world scenarios~\cite{fu2022hint,chen2024uncertainty}. EHR-based clinical trial outcome prediction models typically suffer from two key problems: \textbf{(1) Data Gap:} the disconnect between the data and its related realistic background knowledge makes it hard to leverage the supportive knowledge in prediction. \textbf{(2) Data Inconsistency:} discrepancies between data from different sources, presenting different data distributions, different patterns of data missing, and different data recording formats, undermine the reliability of prediction results across datasets.

Digital twin creation technology has exhibited great potential in clinical practice~\cite{chaudhuri2023predictive,coorey2022health}, simulating patient physiological state~\cite{herrgaardh2022digital,cappon2023replaybg}, disease progression~\cite{yankeelov2024designing}, and treatment effects~\cite{allen2021digital,Chang2023Understanding}. These technologies leverage historical and real-time data to provide medical professionals with more accurate and comprehensive personalized health information, introducing opportunities in outcome prediction and analysis. Inspired by this, Das~\textit{et al}.~\cite{das2023twin} designed a simple model for cancer patients' EHR digital twin creation, which promotes privacy protection and the outcome prediction in clinical trial simulation. However, this method only relies on the existing datasets when creating digital twins, which fails to leverage knowledge beyond the dataset and limits the accuracy of clinical trial outcome prediction. Traditional digital twin models still face challenges in achieving accurate clinical trial outcome predictions, failing to effectively address data gaps and data inconsistencies. Data gaps are particularly prominent in trial predictions due to variations in ICD codes for different medications and diseases, necessitating the acquisition of real-world background knowledge by the digital twin for clinical trials. Moreover, the data inconsistency problem is greatly exacerbated in EHR data for clinical trial scenarios, posing a challenge to digital twin creation, since EHRs for clinical trials often originate from different sites and thus present greater inconsistencies.

To address these problems, in this paper, we propose a novel approach based on LLMs that creates digital twins to enhance the effectiveness and accuracy of clinical trial outcome prediction.
Often referred to as `world models', LLMs possess not only extensive language understanding capabilities but also a comprehensive understanding of world knowledge, including medical knowledge. This makes LLMs capable of overcoming the aforementioned data gap and inconsistency problems.
Our proposed innovative model \texttt{TWIN-GPT} is fine-tuned on a pre-trained LLM (ChatGPT~\cite{ChatGPT}) on clinical trial datasets, so as to generate personalized digital twins for different patients. 
According to the virtual personalized patient data (i.e., digital twins) generated by \texttt{TWIN-GPT} based on historical and real-time examination results, we find that patients' physiological and pathological states over the clinical trial duration can be accurately forecasted, thereby achieving more precise trial outcome predictions. Moreover, experiments on real-world datasets demonstrate that our \texttt{TWIN-GPT} can account for individual patient variations and disease complexities, producing data that closely aligns with diverse real-world scenarios. It resolves challenges of data gap and inconsistency in EHR while ensuring the protection of patient privacy. Our main contributions are summarized as follows:

\begin{itemize}
\item[$\bullet$] \textbf{Innovative Integration of LLM for Digital Twin Creation:} To the best of our knowledge, we are the first to integrate LLM into digital twin creation and perform knowledge association across datasets, which effectively imputes missing EHR data and provides a more personalized patient modeling approaches, thus addressing the limitations of traditional models.
\item[$\bullet$] \textbf{Enhanced Personalization and Accuracy:} 
\texttt{TWIN-GPT} harnesses the vast medical knowledge embedded within ChatGPT to generate personalized digital twin models for individual patients. Experimental findings showcase that this approach achieves significantly enhanced personalization by accounting for each patient's unique characteristics and disease complexities, thereby improving the accuracy of predicted clinical trial outcomes.
\item[$\bullet$] \textbf{Privacy Protection and Versatility in Application:} Our \texttt{TWIN-GPT} approach also protects the patient privacy by generating virtual patient data and simulating personalized physiological measures over time, minimizing the use of sensitive patient information. The versatility allows for application across various clinical trial scenarios, thereby accelerating the pace of clinical trials and enhancing medical research and patient care~\cite{das2023twin}. 
\end{itemize}

\section{RELATED WORK}

\subsection{Large Language Models (LLMs) for Medicine}
LLMs have exhibited excellent performance on various tasks, especially in the field of medicine~\cite{thirunavukarasu2023large,clusmann2023future,karabacak2023embracing}. 
The current research on LLMs in the medical field has provided novel applications in bedside diagnosis, automated drafting of clinical documents, and improving medical workflow efficiency~\cite{cascella2023evaluating,peng2024model,nori2023capabilities}. LLMs have demonstrated their utility in clinical applications by passing medical licensing exams and offering superior empathy and quality of medical advice compared to human clinicians~\cite{Wang2023Augmenting,huang2023chatgpt,oh2023llm,tan2024fine,wu2023medical,yan2024making,yan2024serval}. In clinical applications, LLMs were used to facilitate the analysis of large amounts of textual data, providing insights that can drive medical discoveries and increase the efficiency of research workflows~\cite{waisberg2023large,wang2024jmlr,shi2024ehragent,ban2023query}. They also have significant advantages in daily clinical tasks like clinical documentation drafting, and can significantly reduce the administrative burden on medical staff~\cite{yuan2024continued,goyal2024healai,cao2023llm,fleming2024medalign}. However, there is no research using LLMs in clinical trial scenarios, which is a crucial area demanding the extensive medical knowledge encoded within these large language models.


\subsection{Patient Outcome Prediction}
Previous research in patient outcome prediction primarily relied on traditional machine learning methods such as decision trees, support vector machines, and logistic regression~\cite{ching2018opportunities, hassanipour2019comparison,chang2019machine,chen2020doctor}. These approaches often utilized manually selected features from high-dimensional clinical data. However, manual feature selection frequently fails to incorporate informative features, thereby failing to fully capture individualized patient information and complex data relationships~\cite{choi2017generating}. Recently, deep learning methods have gained prominence with the advancement of deep learning technology. Researchers have begun to explore the use of neural networks for patient prediction~\cite{choi2016doctor,ren2019deep,choi2017using,kam2017learning,chen2024congenital,chen2023excelformer}, which handle well with large-scale data and automatically select informative features, leading to significant achievements. Fu~\textit{et al.}~\cite{fu2022hint,fu2023automated} proposed a hierarchical interaction network (HINT) to predict the clinical trial outcome based on drug molecules, disease codes, eligibility criteria, etc. 
Lu~\textit{et al.}~\cite{lu2024uncertainty} enhance HINT from the perspective of uncertainty quantification and interpretability. 

Among these, analysis of EHR data plays a crucial role in patient outcome prediction. Many studies have investigated how to effectively utilize EHR data for patient prediction~\cite{ching2018opportunities,gupta2022obesity,rajkomar2018scalable,yan2024serval}, applying long-term historical EHR learning, medical text mining, time-series data analysis~\cite{jensen2012mining,amirahmadi2023deep,zhang2021feature}. Utilizing EHR data can provide comprehensive patient information, but it also suffer from data gap problem and data inconsistency problem. Recently, digital twin creation has emerged as a noteworthy field in EHR learning, offering new opportunity to patient outcome prediction~\cite{vallee2023digital}. Digital twin models can simulate a patient's personalized physiological states, disease progression, and treatment effects based on historical EHR and real-time data~\cite{das2023twin}.

\subsection{EHR Data Generation}

Data generation methods can create synthetic data that is effective in addressing data scarcity issues. EHR data generation methods often synthesize patient records conditioned on a patient's historical data and distribution information, using techniques such as generative adversarial networks (GANs)~\cite{baowaly2019synthesizing, choi2017generating,chen2022me} or variational auto-encoders (VAEs)~\cite{allen2021digital, biswal2021eva,chen2021electrocardio,fu2021mimosa,fu2022reinforced}. The advantage of these methods is their ability to protect patient privacy, but ensuring the generated data maintains consistency with real data~\cite{saxena2021generative} remains a challenge.

To protect patient privacy, many approaches employ data masking and anonymization techniques in learning EHR data~\cite{fung2010privacy,yoon2023ehr,venugopal2022privacy,wang2024igamt}. These methods involve de-identifying or anonymizing sensitive information to create anonymous or partially anonymous data. While these methods provide privacy protection, they may potentially lead to a decrease in data quality~\cite{dwork2013toward, brickell2008cost}. Model-driven generation methods use known patient models or epidemiological knowledge to generate synthetic EHR data. These models can take into account a patient's physiological states, disease progression, and treatment effects, resulting in more accurate data generation~\cite{zhang2021synteg,cannesson2019machine,chu2020treatment}. Typically, these methods require domain experts' knowledge to guide the generation process. Some data augmentation techniques can also increase data diversity and quantity using existing EHR data. This includes extracting information from medical texts using natural language processing techniques or creating new data points by combining different types of medical events~\cite{che2017boosting,yahi2017generative, esteban2017real,yan2023text2tree}. Data augmentation can enhance the training effectiveness of models but is still subject to limitations imposed by the original data.

Digital twin is a specialized method for generating personalized EHR data, ensuring similarity and correspondence with the original data, as well as guaranteeing patient privacy protection~\cite{lu2023machine}. Das~\textit{et al}.~\cite{das2023twin} used a VAE to model digital twins in clinical trials to achieve prediction of patient outcome in low-data scenarios. Liu~\textit{et al}.~\cite{liu2019novel} proposed a cloud healthcare system framework based on digital twin healthcare (CloudDTH), which is used to monitor, diagnose, and predict personal health to achieve the goal of personal health management. Zhong~\textit{et al}.~\cite{zhong2022multidisciplinary} used ICU EHR data to build the ICU digital twin model to study critical care services in the ICU. However, these approaches only perform digital twinning within their datasets and do not solve the data gap and data inconsistency problems.

\section{METHODOLOGY}

In this section, we introduce \texttt{TWIN-GPT}, a novel approach that creates personalized digital twins for effective virtual clinical trial simulation, leveraging the power of large language models (LLMs) to address the challenges of clinical trials in limited data scenarios. Our method originates from the understanding that clinical trials are essential for medical research and drug development, but they are often hindered by the extensive time and participant involvement required. \texttt{TWIN-GPT} aims to revolutionize this by generating high-fidelity, personalized digital twins of patients, thus facilitating virtual clinical trials that can significantly enhance patient safety, expedite development, and reduce costs.

\noindent\textbf{Overview.} 
In Section~\ref{sec:formulation}, we define the problem of digital twin creation for clinical trials. Section \ref{sec:prompt}, We employ a novel prompt-tuning approach to generate and update patient digital twins based on their EHR data, enhancing the model's predictive accuracy and understanding of patient outcomes. In Section \ref{sec:model details}, We propose \texttt{TWIN-GPT} to generate digital twins of patients in clinical trials, enhancing prediction accuracy and understanding of patient outcomes by leveraging historical and similar patient data.

\subsection{Problem Formulation and Overview Pipeline}
\label{sec:formulation}

We assume that there are a total of $N$ participant patients in a clinical trial, and each patient $n$'s trial encounters are encapsulated as
\begin{equation}
    X_{n;1:T_n} = \left\{x_{n,1},x_{n,2},\cdots,x_{n,T_n}\right\}, 
\end{equation}
where each $x_{n,t}$ is a collection of event types occurring at the $t$-th visit, defined as
\begin{equation}
x_{n,t} = \left\{x_{n,t}^1,x_{n,t}^2,\cdots,x_{n,t}^u\right\}. 
\end{equation}
Each element $x^i_{n,t}$ corresponds to a kind of clinical event like a treatment or an adverse reaction, and $u$ is the total number of all types of events, while
$x_{n,t}^i = \left\{c_1,c_2,\cdots,c_n\right\}$ represents the occurrence of event, where $c_l$ is 1 or 0.
Formally, we synthesize the digital twin of a patient by an auto-regressive model $M$, and each step predicts the next visit event:
\begin{equation}
\hat{X}_{n,t} = M\left(X_{n^{(')};1:t-1
}, \{X_{n';1:t-1}\}_{n'=1}^{N'}, \Theta\right),
\end{equation}
where $t=1,\ldots,T_n$, and $\Theta$ denotes the model parameters. The term $N'$ is the count of nearest neighbor patients for reference in the digital twin generation. In the training process, we supervise the model $M$'s learning by comparing the generated digital twins with the true patient's trial encounters during the trial. The learning objective is to maximize the cosine similarity between real and synthetic patient representations, 
\begin{equation}
\label{eqn:loss}
\underset{\Theta}{\arg\max}\ \frac{1}{T_n} \sum_{t=1}^{T_n} {sim}(F_{n,t},\hat{F}_{n,t} ),
\end{equation}
\noindent where ${F}_{n,t} $ and $\hat{F}_{n,t}$ denote the average representations of $x_{n,t}^u$ and $\hat{x}_{n,t}^u$ (where $\hat{x}_{n,t}^u$ is in $\hat{X}_{n;1
}$) over the time span from $1$ to $T$. We use cosine similarity as the similarity metric (i.e., ``\textit{sim}'' in Eq.~\ref{eqn:loss}). 

In our analysis, we particularly focus on three essential clinical events (i.e., $u$): 
\begin{itemize}
\item \textit{treatment}, $x_{n,t}^{\text{treat}}$, is the assigned treatment at the $t$-th timestep for the patient $n$; 
\item \textit{medication}, \(x_{n,t}^{\text{med}}\), is defined as the treatment (i.e., drugs) given to patients in response to their current health conditions;
\item \textit{adverse event}, \(x_{n,t}^{\text{ae}}\), refers to all unexpected health incidents that occur during patient visits. 
\end{itemize}
To predict adverse events in the next visit, denoted as $X_{n,t+1}^{\text{ae}}$, we feed all three event types into the model as the \textit{causality} among these event types is crucial for accurate prediction. 
For example, the doctor might provide medication \textit{Docetaxel} to a patient at the timestep $t+1$ due to the adverse events observed in the last visit. Meanwhile, the adverse event \textit{pain} might not be observed at timestep $t+1$ due to the medication \textit{Docetaxel} given at timestep $t$. 
Since our model is a kind of large language model, the input three event types serve as in-context learning ability~\cite{min2022rethinking} in our case.

\subsection{Prompt Tuning for Digital Twin in LLMs}
\label{sec:prompt}
In this work, we develop a large language model (LLM) to serve as the auto-regressive model $M$ to perform the digital twin creation. To LLMs, a \textit{prompt} is the initial input or query given to the model for generating a better response~\cite{zhao2023survey}. 
A context-suited prompt is beneficial to our digital twin creation task. 
Compared with expensive model tuning, prompt tuning is cheaper. 
We proposed a novel prompt-tuning approach to seek suitable prompts to generate better digital twins. The steps of our proposed \textit{prompt tuning} are as follows:

\begin{itemize}
\item Initially, the first version of digital twin is generated based on the patient's EHR data. At this stage, the patient's EHR data of previous visits and the five most similar EHR data are provided as reference.
\item Next, the real patient records (e.g., presenting the adverse event) are regarded as the ground truth to provide supervision.
We use initial prompts for predicting adverse events and medications. Based on the ground truth data, the LLM's parameters are updated. The \texttt{TWIN-GPT} is developed on the ChatGPT, and we utilize ChatGPT's fine-tuning API to update it.
\item If the correct adverse event under the current treatment is predicted, the medication for the next time is deduced sequentially.
\end{itemize}

Steps 1-3 are repeated each time the patient's EHR data is updated. When the updated EHR data is provided, along with another five most similar EHR profiles, the adverse events are predicted and medication is recommended again, as illustrated in table \ref{table:prompt_tuning} below. The goal of this approach is to track and update the patient's EHR digital twin in real time, obtaining prompts that most easily establish knowledge associations between input and output.

\begin{table}[h]
\centering
\begin{tabular}{ccc}
\midrule
\textbf{Visit} & \textbf{Input} & \textbf{Prompt} \\ \hline
$t$ & 
\begin{tabular}[c]{@{}c@{}}
Target visit: $X_{n;1:t-1}$ \\ 
Five nearest neighbor EHR
\end{tabular} & 
\begin{tabular}[c]{@{}p{10cm}@{}}
Based on the historical visit data of the target patient: xxxxxx, please use the patient's history and the k most similar visits to generate a digital twin of the target patient and predict the digital twin's adverse events.
\end{tabular} \\ \hline
$t+1$ & 
\begin{tabular}[c]{@{}c@{}}
Target visit: $X_{n;1:t}$ EHR \\ 
Five nearest neighbor EHR
\end{tabular} & 
\begin{tabular}[c]{@{}p{10cm}@{}}
Based on the historical visit data of the target patient: xxxxxx, please use the patient's history and the k most similar visits to generate a digital twin of the target patient and predict the digital twin's recommended medications.
\end{tabular} \\ \bottomrule
\end{tabular}
\caption{Illustration of Prompt Tuning for Digital Twin.}
\label{table:prompt_tuning}
\end{table}

\subsection{Digital Twin Generation}
\label{sec:model details}
In clinical trials, the sequence of real patient visit records is defined as $X_{n;1:T_n}$. Our goal is to generate a digital twin $\hat{X}_{n;1:T_n}$ that can retain the features of the target patient. For this purpose, we propose a generator, namely \texttt{TWIN-GPT}, that simulates personalized patient trajectories based on the real record $X_{n;1:T_n}$. 
The \texttt{TWIN-GPT} predicts in an auto-regressive manner, 
\begin{equation}
\hat{X}_{n;T_n-1:T_n} = \texttt{TWIN-GPT}(X'_{n;1:T_n-1} , \{X_{k;1:T_k-1}\}_{k}). 
\end{equation}

In training, $X'_{n;1:T_n-1}$ is $X_{n;1:T_n-1}$, while in test $X'_{n;1:T_n-1}$ is the previous prediction outcomes $\hat{X}_{n;1:T_n-1}$.
To enable \texttt{TWIN-GPT} to learn personalized clinical trial data for patients and to leverage data association between different patients, we adopted the method of prompt tuning introduced above. In this way, we hope \texttt{TWIN-GPT} can accurately generate digital twin characteristics of the patient and finally realize the prediction of different clinical trials. We fine-tune \texttt{TWIN-GPT} following the step described in Sec.~\ref{sec:prompt}, letting it make predictions based on the patient's historical visits so that \texttt{TWIN-GPT} can learn the patient's unique characteristics. Specifically, based on the medication (\(Med\)) and treatment before and planned at the current time point \(t\), the Adverse Event (\(AE\)) at the next time point \(t+1\) is predicted. Then based on the \(AE_{t+1}\), the \(Med_{t+1}\) can be further predicted. Here, the patient's unique characteristics and information from $k$ similar patients are used as references, to ensure the reliability of the digital twins. Please note that in clinical trials, AEs and medication information are crucial factors we need to focus on, as they significantly impact the success and conclusions of the clinical study. Therefore, our \texttt{TWIN-GPT} is primarily designed to predict these two events.

\begin{figure*}
  \centering
  \includegraphics[width=0.8\textwidth]{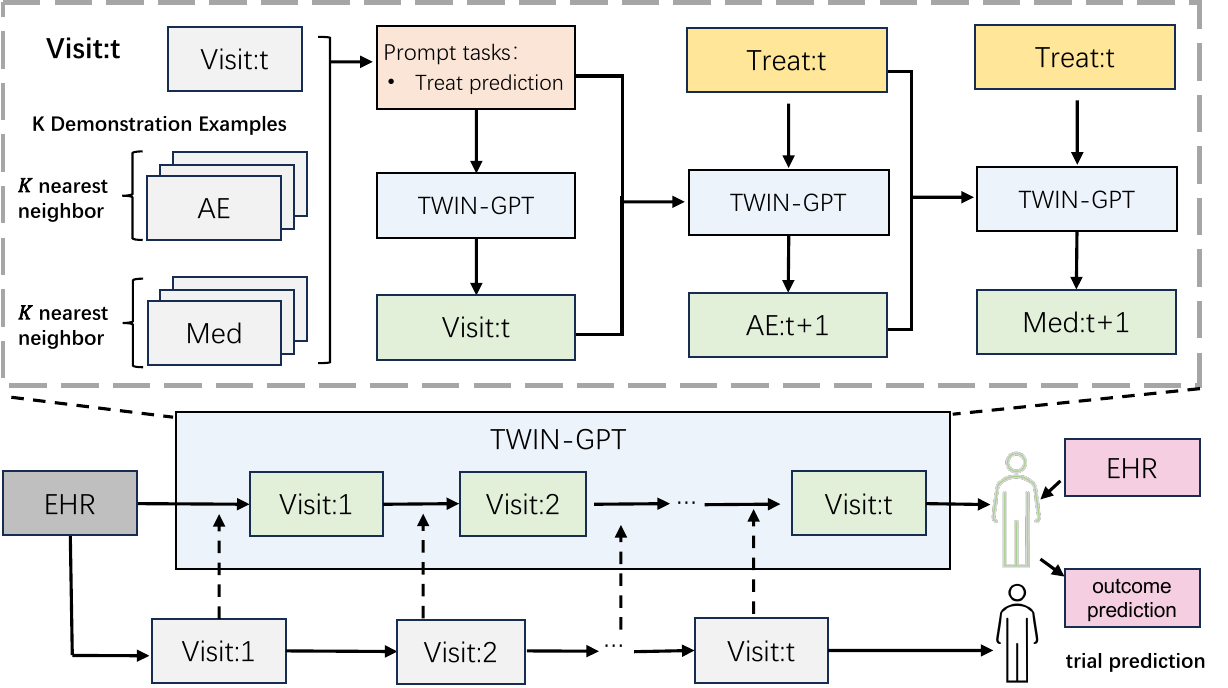}
  \caption{The workflow of \texttt{TWIN-GPT}. (Bottom) \texttt{TWIN-GPT} takes real follow-up visits $X_{n,1:T_n-1}$ of a patient and generates twin visits of next step, $\hat{X}_{n,T_n}$. Finally, the whole visit sequence can be predicted. (Top) The top part elaborates on how to use the digital twin visits $\hat{x}_{n,1:t}$ (at the time step $t$) to predict the events that occurred in the next timestamp $\hat{x}^{\text{event}}_{n,t+1}$. We use \textit{K} nearest neighboring patient visits in \texttt{TWIN-GPT} fine-tuning but only use origin visits in prediction. ``Treat'': treatment; ``Med'': medication; ``AE'': adverse event.}
  \label{fig:framework}
\end{figure*}

Notably, different patients may have different trajectories in the trial. To enable \texttt{TWIN-GPT} to create digital twins for various patient patterns, the $k$ most similar patients are retrieved based on all of their previous event sequential $X_{1:t}$. This approach not only helps improve \texttt{TWIN-GPT}'s comprehensive understanding of patient patterns but also prevents it from overfitting to specific individuals. Specifically, at the time step of the visit $x_{n,t}$, we enumerate all the other patients and retrieve $K$ visits $\left\{x_{k,t_k}\right\}^K_{k=1}$ with the highest cosine similarity to $x_{n,t}$. Note that $t_k$ can be either equal or unequal to $t$. By integrating these similar visits as inputs, \texttt{TWIN-GPT} can also enhance its ability to differentiate between similar visits in prediction. The overall workflow is illustrated in Fig.~\ref{fig:framework}.

\section{APPLICATION \& EVALUATION}
The generated digital twins are beneficial for simulating clinical trials in silico. In this section, we present two application scenarios (personalized generation and counterfactual generation) and evaluate the quality of the digital twins based on clinical trial outcome prediction accuracy and privacy protection.

\subsection{Digital Twins Generation Application}

\subsubsection{Personalized Generation}

Personalized Generation refers to the process of using EHR data to guide TWIN-GPT in replicating patients, ultimately generating digital twins.
Throughout each step of this process, the model incorporates both its data and data from the $k$-nearest neighbors, enabling \texttt{TWIN-GPT} to perform the in-context learning.

\subsubsection{Counterfactual Generation}
The purpose of this task is to simulate the trajectories of patients under alternative treatment schedules, including switching a patient from the treatment arm (\textit{T}) to the control arm (\textit{C}). This simulation not only enhances patient records but also enables the estimation of personalized treatment effects.  Moreover, it facilitates the balancing of trial data for predictive modeling while substantially reducing the necessary sample size for recruiting participants in the control arm.
To generate the counterfactual digital twin $\hat{X}_{n,1:T_n}$, corresponding to the real patient record $X_{n,1:T_n}$, we need to identify the most closely matching patient record within  $\tilde{X}_{k,1:T_k} $, where  $n$ is among $T$ and $k$ belongs to $C$. This step merges the unique personal traits of patient $n$ with the temporal patterns observed in patient $k$, facilitating the creation of the synthetic pathway. We evaluate the applicability of \texttt{TWIN-GPT} trajectory generation by comparing the similarity between the synthesized trajectory and the most similar patient record.

\subsection{Clinical Trial Outcome Prediction Evaluation of Digital Twins }

To assess the practicality of digital twin models, we observe whether the additional synthetic data generated by models can successfully predict outcomes under different treatment modalities. Specifically, our experiments involve the following prediction tasks:

\subsubsection{Dimension-wise probability.}
To evaluate the similarity between twin data and the real data, we perform a dimensional-wise probability calculation. The calculation of dimension probabilities helps us to understand the probability of occurrence of each feature in the dataset, and it can provide quantitative metrics about the consistency and similarity between synthetic and real data. The Dimension-wise probability (DP) is calculated by 
\begin{equation}
DP = \frac{V_f}{V_l},
\end{equation}
where $V_f$ is the number of visits in the dataset containing a specific feature. $V_l$ is the total number of accesses in the dataset.
When digital twin models successfully produce synthetic data that closely mirrors the original, the distributional properties (DPs) of the synthetic data will approximate those of the real data. We calculate the Person correlation coefficient, $r \in [0,1]$. If $r = 1$, it means that the model has high fidelity.

\subsubsection{Counterfactual digital twin evaluation.} In the evaluation of counterfactual digital twins, for a specific patient, we identify the closest match from different treatment groups to act as a substitute for the inaccessible counterfactual outcomes, in line with established practices in causal inference research~\cite{shalit2017estimating}. We assess the similarity between the digital twin and its corresponding record by calculating personalized Pearson correlation coefficients and quantify the fidelity by comparing data points between the synthetic and surrogate records.

\subsubsection{Severe outcome prediction}

To validate whether digital twin models can fully comprehend and replicate the underlying relationship between real-world data records and severe outcomes, we conducted severe outcome prediction validation. Firstly, we defined severe outcomes to include deaths and other critical clinical events. Then, we employed Long Short-Term Memory (LSTM) to forecast these severe outcomes. LSTM received real clinical data and synthetic data generated by different models for model training and prediction. Ultimately, we utilized the Area Under the Receiver Operating Characteristic Curve (AUROC) to assess the accuracy and performance of the predictions.

\subsubsection{Adverse event prediction}
Adverse event prediction is to show how much causal relation the synthetic data generated by digital twin models have maintained compared to the raw clinical data. We train another MLP (multiple-layer perceptron) to evaluate the prediction performance by inputting synthetic data and real data separately. 

\subsection{Digital Twins Clinical Privacy Evaluation}

In the generation and utilization of synthetic data, privacy protection is of paramount importance. This section will delve into three primary methods of privacy risk evaluation: Presence Disclosure, Attribute Disclosure, and Nearest Neighbor Adversarial Accuracy Risk (NNAA).

\subsubsection{Presence Disclosure}


To evaluate the risk of privacy leakage, where an attacker analyzes synthetic data to infer whether the records of a particular individual participated in the training of the model, we employ sensitivity as a crucial metric.  Sensitivity serves as an important indicator for assessing the privacy risk associated with synthetic data.  By comprehending sensitivity, we can gain a clearer understanding of the level of privacy protection offered by synthetic data and implement appropriate measures to mitigate the risk of existential leakage.  The sensitivity is calculated using the following formula: 
\begin{equation}
S = \frac{R_k}{R_l}, 
\end{equation}
where $R_k$ is the number of known records found in the synthetic data and $R_l$ is the total number of known records. For instance, if we have 100 known records and detect 20 of them in the synthetic data, the sensitivity would be $20/100=0.2$. This implies a 20\% probability for an attacker to confirm the presence of known samples in the training data. By calculating the sensitivity, we can determine the likelihood of an attacker identifying a known sample within the training data.  If the synthetic data contains a limited number of known records, resulting in low sensitivity, it indicates a relatively secure nature of the synthetic data with a reduced risk of existential leakage.  Conversely, if the synthetic data contains a substantial number of known records, leading to a high sensitivity, the risk of existential leakage becomes more pronounced.

\subsubsection{Attribute Disclosure}

Privacy leakage is when an attacker is able to obtain or infer sensitive information or unknown attributes of an individual, thereby violating an individual's right to privacy. Attribute leakage occurs when the attacker is able to infer other attributes of the patient based on a known subset of the data. To quantify this risk, we adopt average sensitivity as a metric. Average sensitivity measures the attacker's ability to infer unknown attributes.  Mean sensitivity(MS) is calculated using the following formula: 


\begin{equation}
MS = \frac{1}{N} \sum_{v=1}^{N} \frac{F_d}{F_l},
\end{equation}
where $F_d$ represents the number of unknown features discovered, and $F_l$ represents the total number of unknown features.   For example, if an attacker knows ten attributes of a patient and can infer two unknown attributes from the synthetic data, the average sensitivity would be $2/10=0.2$. This implies that the attacker has a 20\% probability of confirming the presence of these unknown attributes in the synthetic data.
By computing the average sensitivity, we gain insight into the attacker's average capability to infer the patient's unknown attributes. A higher average sensitivity indicates a greater risk of privacy leakage, as it suggests that the attacker is more likely to infer unknown attributes. Conversely, a lower average sensitivity indicates a lower risk of privacy leakage, as it implies that the attacker's inference capability is relatively weak.
\subsubsection{Nearest Neighbor Adversarial Accuracy Risk (NNAA)}

NNAA is a measure of the degree to which a model overfits the original data, directly relating to privacy leakage risk. The NNAA risk score is calculated as follows:

\begin{equation}
\text{NNAA risk score} = AA_{ES} - AA_{TS}, 
\end{equation}
where $AA_{ES}$ and $AA_{TS}$ represent the aggregated distances between synthetic data and evaluation data, and between synthetic data and real data, respectively. They are calculated as follows:

\begin{equation}
\resizebox{.5\textwidth}{!}{$AA_{ES} = \frac{1}{2} \left( \frac{1}{N} \sum_{i=1}^{N} 1(d_{ES}(i) > d_{EE}(i)) + \frac{1}{N} \sum_{i=1}^{N} 1(d_{SE}(i) > d_{SS}(i)) \right)$}. 
\end{equation}


\begin{equation}
\resizebox{.5\textwidth}{!}{$AA_{TS} = \frac{1}{2} \left( \frac{1}{N} \sum_{i=1}^{N} 1(d_{TS}(i) > d_{TT}(i)) + \frac{1}{N} \sum_{i=1}^{N} 1(d_{ST}(i) > d_{SS}(i)) \right)$}. 
\end{equation}

Here, $d_{ES}(i)$, $d_{TS}$, $d_{SE}$ and $d_{EE}$, $d_{TT}$, $d_{SS}$ represent the nearest neighbor distances among synthetic data, evaluation data, and real data respectively. For example, if the average distance between synthetic data and evaluation data is greater than that between synthetic data and real data, the NNAA risk score will be positive, indicating potential over-fitting of the model to the training data, hence increasing the risk of privacy leakage.

Through these three methods, we can more comprehensively evaluate the privacy risks of synthetic data, thereby better protecting individual privacy.

\section{EXPERIMENTS}

\subsection{Experimental Setup}

\subsubsection{Data source}

\noindent \textbf{Original clinical trial dataset:} 
We trained \texttt{TWIN-GPT} using a Phase III breast cancer trial dataset (NCT00174655), analyzing Disease-Free Survival (DFS) across different treatment methods.  The study randomly divided 2,887 patients into groups to compare the efficacy of docetaxel, either alone or with doxorubicin followed by CMF, against doxorubicin alone or combined with cyclophosphamide, then CMF, in patients with positive axillary lymph nodes.  This dataset, publicly available in Project Data, facilitated our performance evaluation.

\noindent \textbf{Trial Outcome Prediction (TOP) dataset:} We also utilized the multi-modal TOP dataset~\cite{fu2023automated} for predicting clinical trial outcomes. This dataset integrates clinical trial-related information from multiple data sources, with each trial record including drug molecule information, disease information, eligibility criteria, and trial result information. The dataset is divided into three phases: 1,160 Phase I trials, 4,449 Phase II trials, and 3,436 Phase III trials. The drug molecule information includes the names of candidate drugs and their functional groups. The disease information includes the ICD-10 codes, disease descriptions, and disease hierarchy represented by CCS codes. The eligibility criteria for the trials are described using unstructured natural language, including inclusion and exclusion criteria. The trial result information includes binary indicators for trial success (1) or failure (0), trial phase, start and end dates, trial sponsor, and trial scale (i.e., the number of participants).

\subsubsection{Data preprocessing}
We processed this dataset by extracting medications, treatments, and adverse events from the original clinical trial data. Then we selected the top 100 most frequently used medications and the top 50 most frequent adverse events, which contain most of the information in this dataset. We also merged all rare events (frequency < 50) into one additional adverse event, so that the model could fully utilize all samples.
Regarding the TOP dataset~\cite{fu2023automated}, we utilized three distinct phases of this dataset to construct a twin network model. The study fully capitalized on the complete information available in the TOP dataset, wherein the criteria for trial qualification were delineated as background knowledge for establishing twin states, whereas the rest was considered clinical medical data for developing twin characteristics.

\subsubsection{Baseline Models}
We compare \texttt{TWIN-GPT} with the following baseline methods: 
\begin{itemize}
\item \textbf{EVA~\cite{biswal2021eva}} utilizes variational autoencoders to create synthetic versions of electronic health records, serving as a generative model for this purpose.
\item \textbf{SynTEG~\cite{zhang2021synteg}} is a representative generative model that employs GANs to produce synthetic versions of EHRs. It uses transformer~\cite{vaswani2017attention} as neural architecture in the dependency extraction part and Wasserstein GAN with gradient penalty (WGAN-GP)~\cite{arjovsky2017wasserstein, gulrajani2017improved} in the conditional generation part.
\item \textbf{PromptEHR~\cite{wang2022promptehr}} leverages real EHRs to train a prompt learning-based generative language model for synthetic EHR generation.
\item \textbf{TWIN-VAE~\cite{das2023twin}} can detail clinical data through the development of individualized clinical trial digital twins, utilizing variational autoencoders (VAE) for this process.  
\item \textbf{$k$-NN-based method~\cite{beigi2022synthetic}} is a simple model that modifies the real patient data by picking random parts from its nearest neighbors.
\end{itemize}

\subsection{Generation Quality}
\subsubsection{Personalized Generation.}
\begin{figure}
    \begin{minipage}[t]{0.47\columnwidth}
        \centering
        \includegraphics[width=1.0\columnwidth]{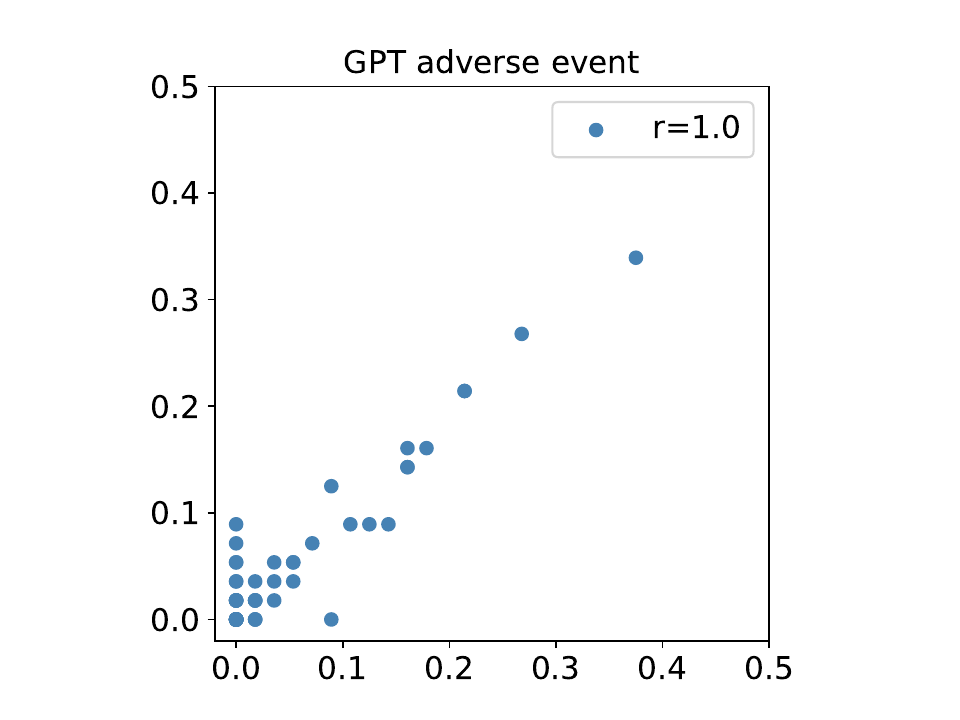}
        \caption{On the \textit{Original clinical trial dataset}, we analyzed the dimension-wise Pearson correlation coefficient (\textit{r}) of adverse events to evaluate the performance of \texttt{TWIN-GPT}. The x-axis displays the probability across dimensions for real data, while the y-axis signifies the probability associated with synthetic data.}
        \label{fig:Fidelity}
    \end{minipage}\hfill
    \begin{minipage}[t]{0.47\columnwidth}
        \centering
        \includegraphics[width=1\columnwidth]{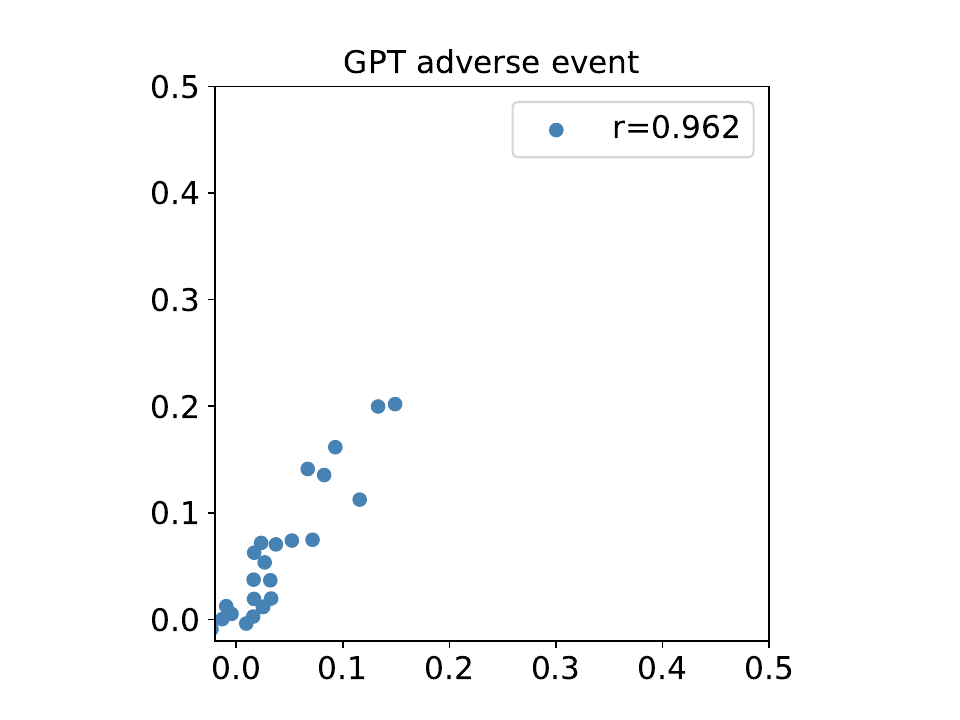}
        \caption{On the \textit{TOP dataset}, we analyzed the dimension-wise Pearson correlation coefficient (\textit{r}) of adverse events to evaluate the performance of \texttt{TWIN-GPT}. The x-axis displays the probability across dimensions for real data, while the y-axis is the probability associated with synthetic data.}
        \label{fig:Fidelity_TOP}
    \end{minipage}\hfill
\end{figure}

\begin{figure}
  \centering
  \subfigure[Phase I]{
  \includegraphics[width=0.31\textwidth]{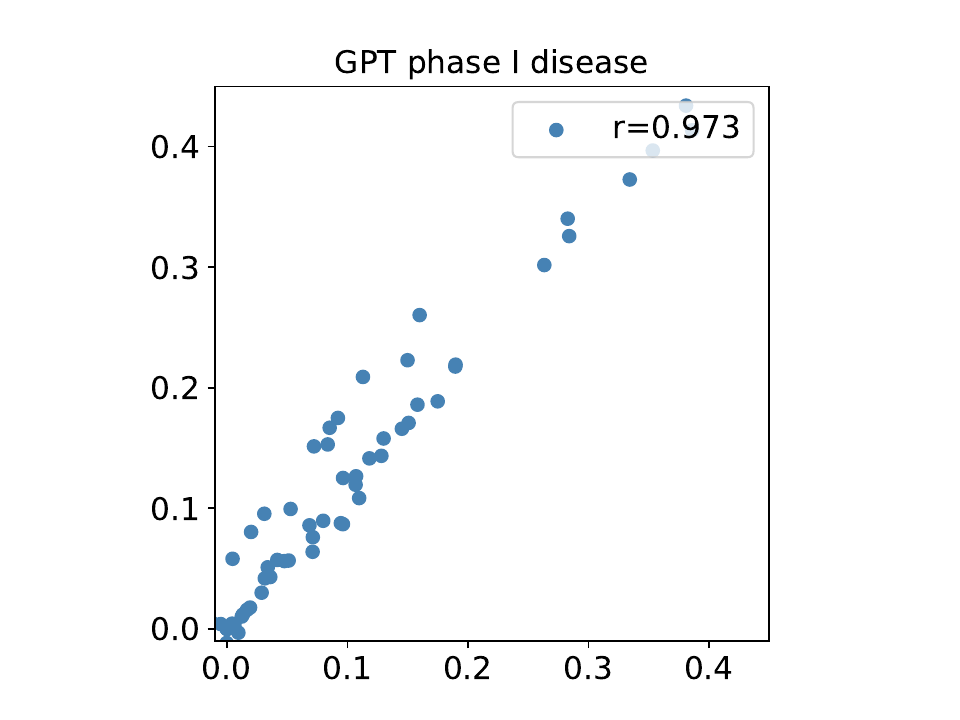}
  }
  \subfigure[Phase II]{
  \includegraphics[width=0.31\textwidth]{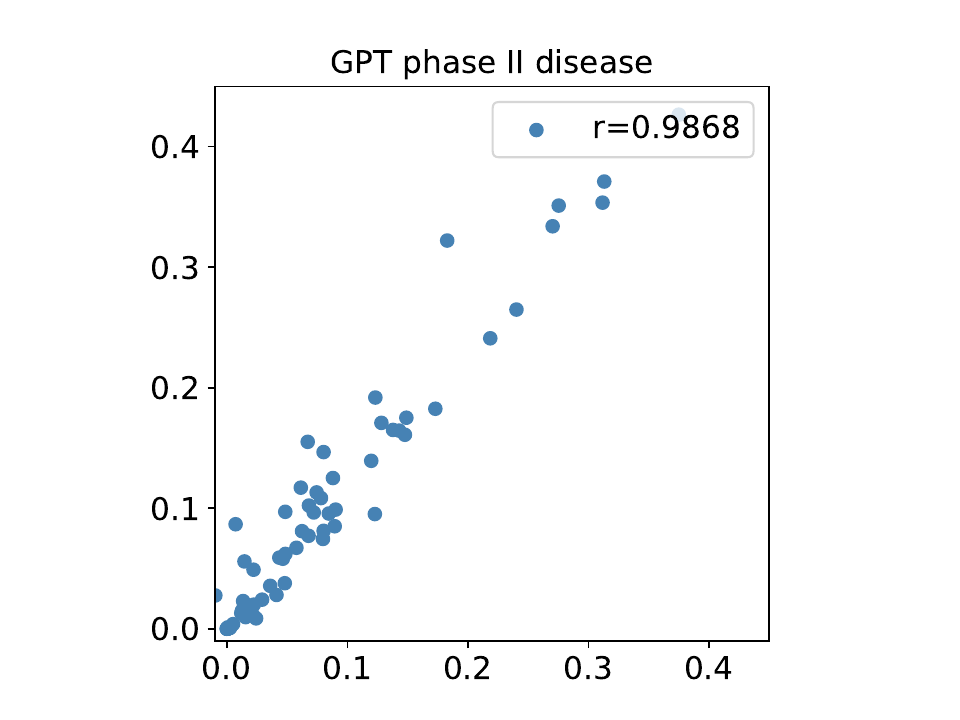}
  }
\subfigure[Phase III]{
  \includegraphics[width=0.31\textwidth]{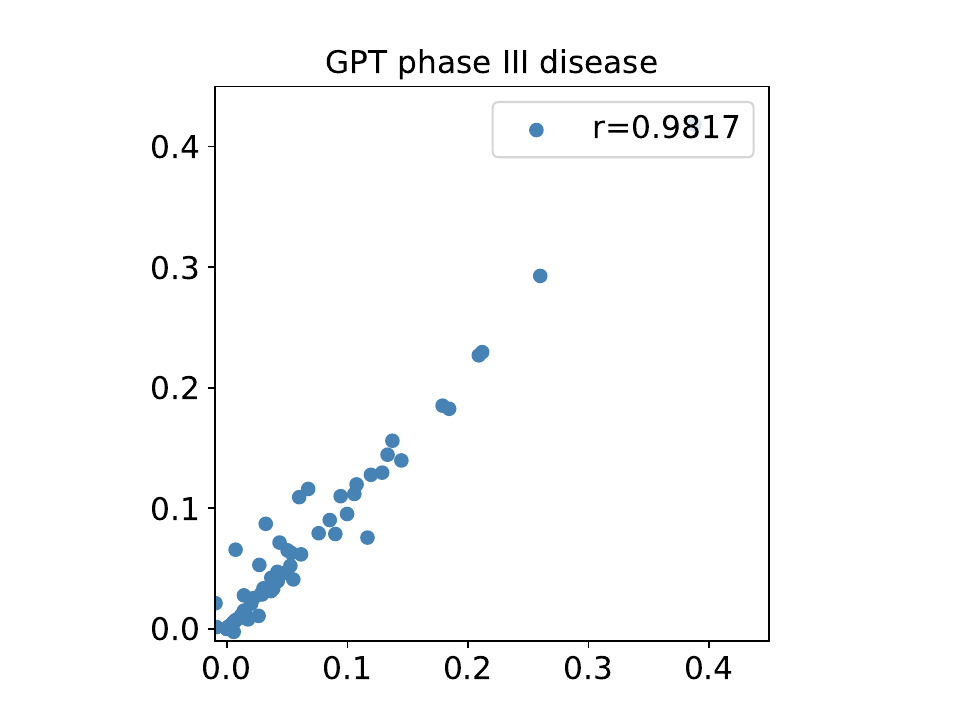}
  }
  \caption{On the \textit{TOP dataset},  we analyzed the dimension-wise Pearson correlation coefficient (\textit{r}) of adverse events in the three phases to evaluate the performance of \texttt{TWIN-GPT}. The x-axis displays the probability across dimensions for real data, while the y-axis signifies the probability associated with synthetic data.}
  \label{fig:Presence_disclosure_TOP_three}
\end{figure}

To make sure that \texttt{TWIN-GPT} has fully learned every dimension's distribution precisely and accurately, we calculate all DPs of every adverse event separately from both real and generated synthetic data. Fig. \ref{fig:Fidelity} shows the performance of  \texttt{TWIN-GPT} for Breast Cancer data. We compare the performance of dimensions' distribution of \texttt{TWIN-GPT} and other baselines. The \textit{r}s of adverse events for \textit{EVA}, \textit{SynTEG}, \textit{PromptEHR}, \textit{$k$-NN}and \textit{TWIN-VAE} are -0.06, -0.07, 0.14, 1.0 and 0.99, respectively~\cite{das2023twin}. We can see that \texttt{TWIN-GPT} has higher fidelity than \textit{SynTEG}, \textit{PromptEHR}, and \textit{TWIN-VAE}. 
Meanwhile, \textit{$k$-NN} achieves performance comparable to \texttt{TWIN-GPT}, which can be easily explained that \textit{$k$-NN} essentially copies and merges fragments from real data to generate synthetic data. However, \textit{$k$-NN} cannot be used to generate EHR data directly because all the generated data can be linked to their original patient's record, which will lead to the privacy risk of leaking information. 
Regarding the TOP dataset, we conducted overall DP calculations for the three medical phases in Fig. \ref{fig:Fidelity_TOP} and independent DPs calculations for the three clinical trial phases as shown in Fig. \ref{fig:Presence_disclosure_TOP_three}. In both scenarios, TWIN-GPT demonstrated high performance with $r\geq$0.96. Since the TOP dataset is a multimodal dataset, the comparison algorithms used in the previous dataset do not apply to this dataset. Here, we used \textit{$k$-NN} to calculate for the \textit{TOP dataset}, where its $r$ is 1, slightly higher than our \texttt{TWIN-GPT}. As mentioned before, \textit{$k$-NN} is not suitable due to the potential for privacy risks associated with linking generated data back to the original patient records.


We also find out per-patient DPs to determine how well each synthetic record matches the associated synthetic record. We compute the Pearson Correlation Coefficient, \textit{r}. The comparison of distributional properties (DPs) between synthetic and real data across all patients is analyzed through Pearson Correlation Coefficients, shown in a histogram within Fig.~\ref{fig:Patient_wise_Pearson}. This comparison indicates the synthetic data's high fidelity on a feature-wise level, with a significant portion of correlation coefficients (\textit{r} values) surpassing 0.8.


\subsubsection{Counterfactual Generation.}
To assess the quality of the counterfactual generation results, we train an LSTM model on real data to predict severe outcomes. Subsequently, we utilize this trained model to predict the counterfactual digital twins generated by \texttt{TWIN-GPT}, representing the simulated patients assigned to the alternative treatment arm. We then compare these predictions with the surrogate ground truth outcomes. Notably, the obtained AUROC score of 0.821, which is quite close to 0.838, AUROC score of the predictions with real data,  demonstrates the high fidelity of the generated counterfactual digital twins.

\begin{figure}
    \begin{minipage}[t]{0.48\columnwidth}
        \centering
        \includegraphics[width=1.0\columnwidth]{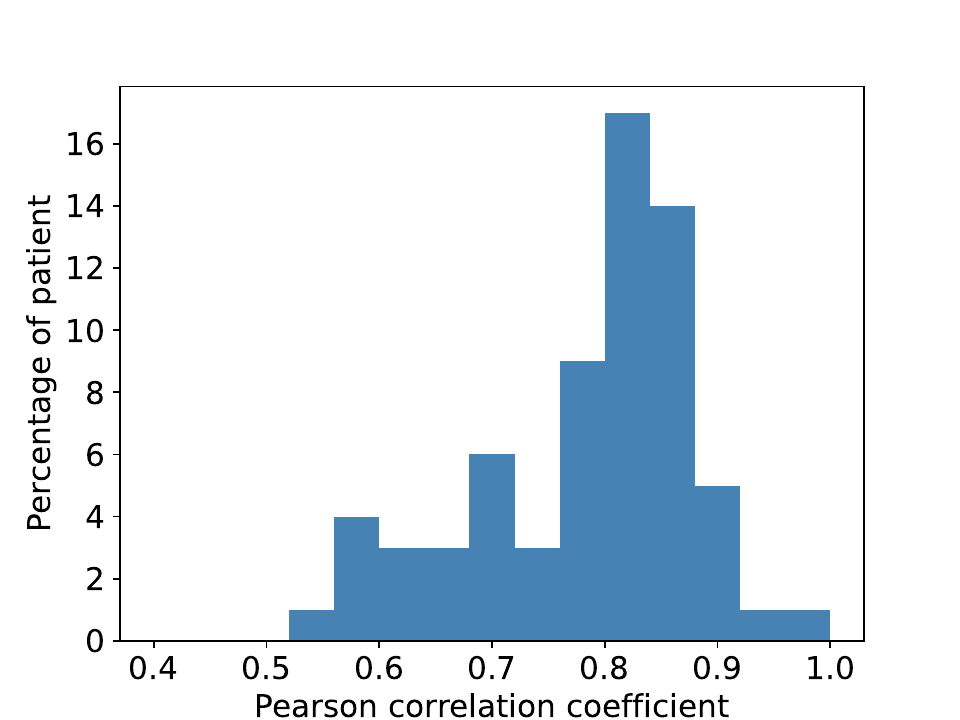}
        \caption{Patient-wise Pearson correlation coefficient (\textit{r}) for \texttt{TWIN-GPT}. \textit{r} is charting the distributional properties (DPs) of each patient's closest match on the x-axis against the DPs of their respective synthetic digital twin on the y-axis. Most of the participants have high fidelity (\textit{r} larger than 0.8).}
        \label{fig:Patient_wise_Pearson}
    \end{minipage}\hfill
    \begin{minipage}[t]{0.48\columnwidth}
        \centering
        \includegraphics[width=1.0\columnwidth]{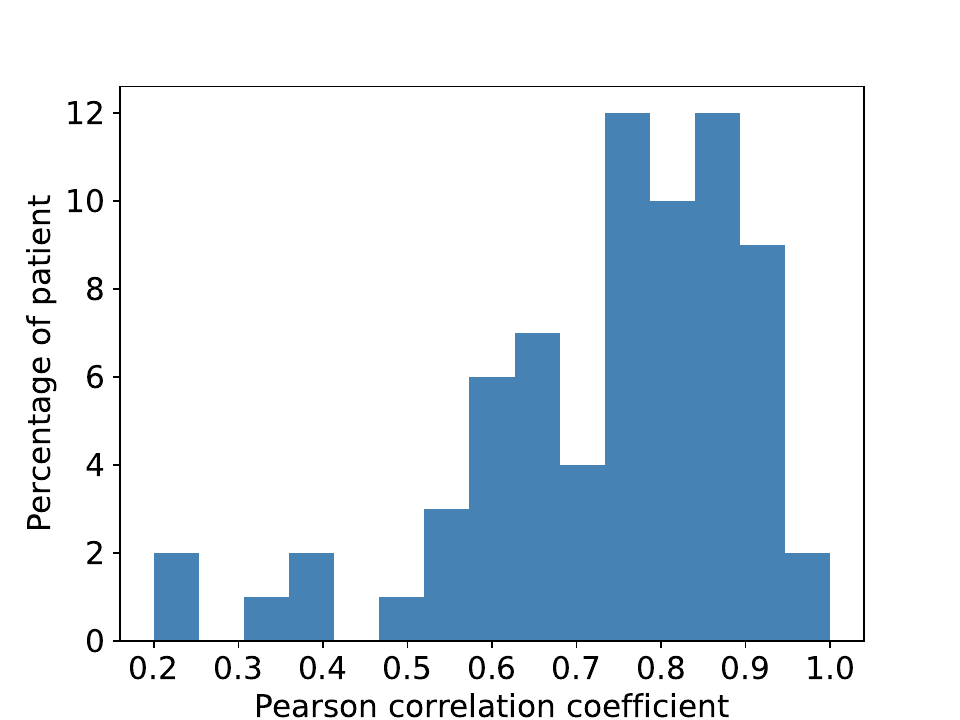}
        \caption{Counterfactual Generation (\textit{r}) for \texttt{TWIN-GPT}. \textit{r} is charting the distributional properties (DPs) of each patient's closest match on the x-axis against the DPs of their respective synthetic digital twin on the y-axis. Most of the participants have high fidelity (\textit{r} larger than 0.8).}
        \label{fig:Counterfactual_Generation}
    \end{minipage}\hfill
\end{figure}

Just like the way we determine the feature-wise fidelity in the \textit{Personalized Generation} part, we assess the distributional properties (DPs) of actual data from the closest neighbor against the DPs of the corresponding synthetic data to determine the Pearson correlation coefficient ($r$). This comparison yields coefficients that are represented in Fig.  \ref{fig:Counterfactual_Generation}. The results indicate that the digital twins of most patients exhibit a high degree of similarity to their real-life counterparts. These digital twins exhibit high $r$s that are larger than 0.8.

\subsection{Digital Twins Clinical Trial Prediction}
\subsubsection{Severe outcome prediction}
\begin{figure}
    \begin{minipage}[t]{0.48\columnwidth}
        \centering
        \includegraphics[width=1.0\columnwidth]{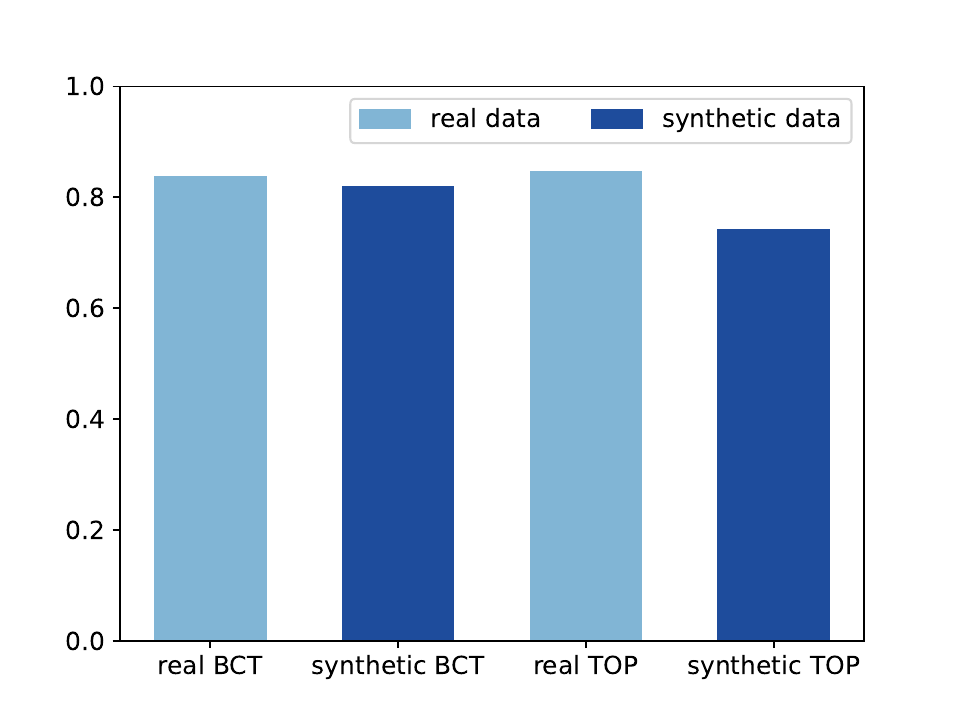}
  \caption{Severe outcome prediction results measured by AUROC scores. The AUROC scores of prediction made by LSTM when taking in real and synthetic data are quite close. }
  \label{outcome prediction}
    \end{minipage}\hfill
    \begin{minipage}[t]{0.48\columnwidth}
        \centering
        \includegraphics[width=1.0\columnwidth]{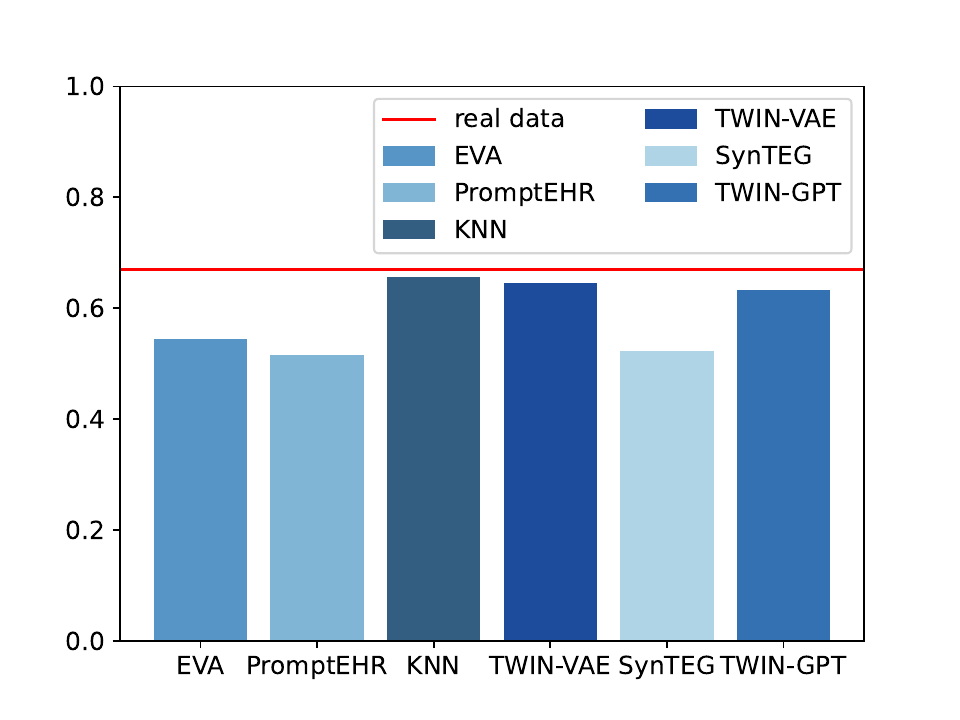}
  \caption{Adverse event prediction AUROC scores. The red line means the AUROC score of MLP to predict the adverse event by taking in real data. We can see that the synthetic data generated by \texttt{TWIN-GPT} and $k$-NN-based model have similar performance when the MLP predictor is trained on synthetic data and real data.}
  \label{Adverse event prediction AUROC scores}
    \end{minipage}\hfill
\end{figure}

We train an LSTM to predict the severe outcome. This model takes in the real clinical data and synthetic data generated by \texttt{TWIN-GPT}. We show this severe outcome prediction result in Fig. \ref{outcome prediction}. The LSTM achieves nearly the same AUROC scores when taking in the synthetic data generated by \texttt{TWIN-GPT} as taking in the real data, indicating that \texttt{TWIN-GPT} can fully understand and reproduce the latent relationship between records and severe outcomes in real data. 

\subsubsection{Adverse event prediction}
This task is to see how much causal relation the synthetic data have maintained compared to the raw clinical data. We train an MLP to predict the adverse event at the next step by taking in real and synthetic data generated by different models separately. Results are shown in Fig. \ref{Adverse event prediction AUROC scores}, where both \texttt{TWIN-GPT} and \textit{$k$-NN-based} model have similar performance with the real data, indicating that our method captures the temporal causal relations accurately. 


\subsection{Digital Twins Clinical Privacy Protection}
Privacy protection is of paramount importance in the generation and utilization of synthetic data. In this part, we evaluate three primary indexes: Presence Disclosure, Attribute Disclosure, and Nearest Neighbor Adversarial Risk (NNAA).

\begin{figure}
    \centering
    \subfigure[Presence disclosure]{
        \includegraphics[width=0.5\columnwidth]{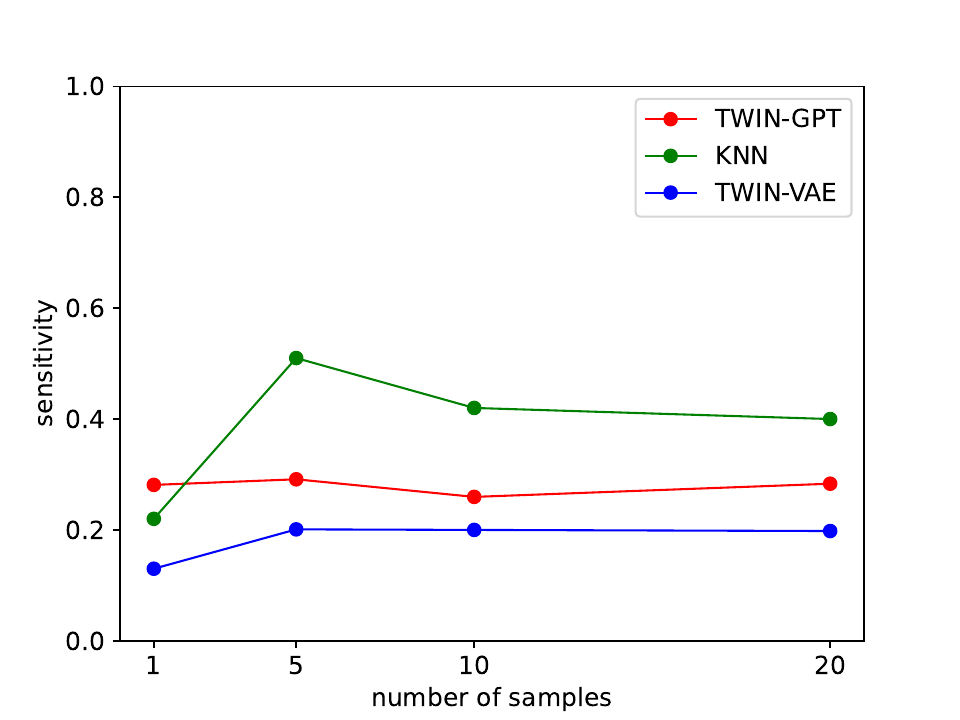}
    }\subfigure[Attribute disclosure]{
        \includegraphics[width=0.5\columnwidth]{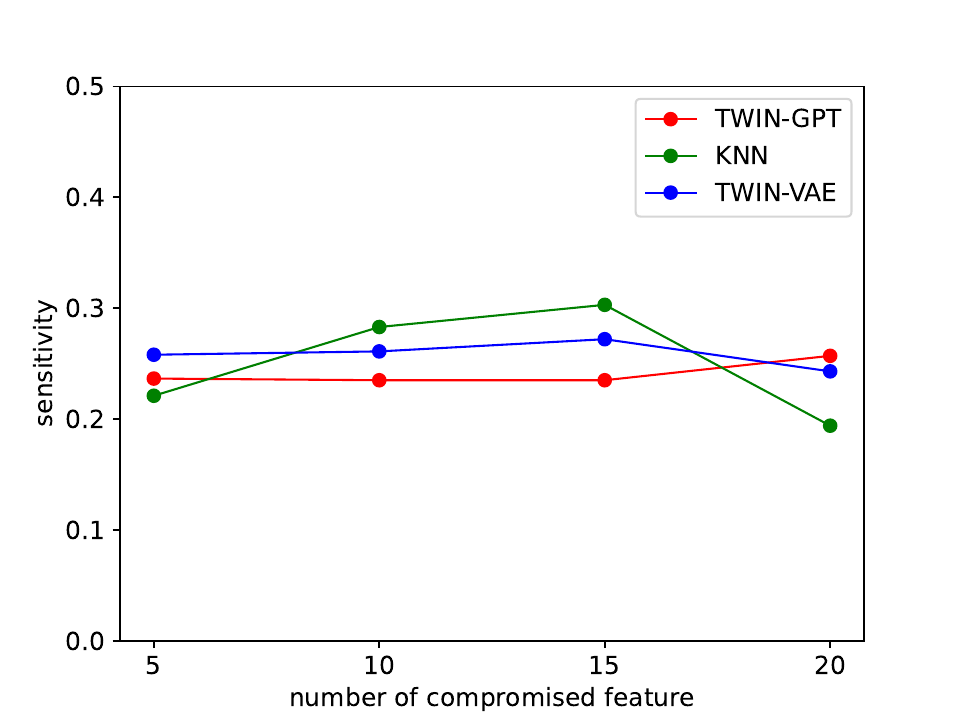}
    }
    \caption{On the \textit{Original clinical trial dataset}, Presence disclosure and Attribute disclosure sensitivity scores with a different number of samples known by the attacker. (Lower sensitivity is better). }
    \label{PandAttribute_oct}
\end{figure}

\begin{figure}
  \centering
  \subfigure[Presence disclosure: comprehensive perspective]{
  \includegraphics[width=0.48\textwidth]{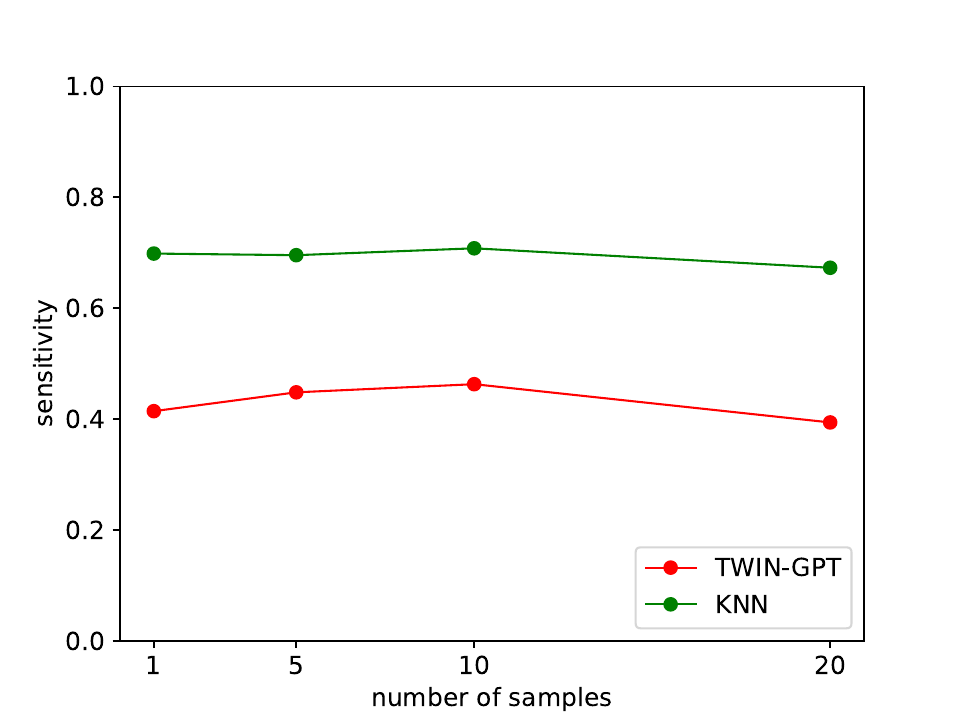}
  }
  \subfigure[Presence disclosure: phase-by-phase analysis]{
  \includegraphics[width=0.48\textwidth]{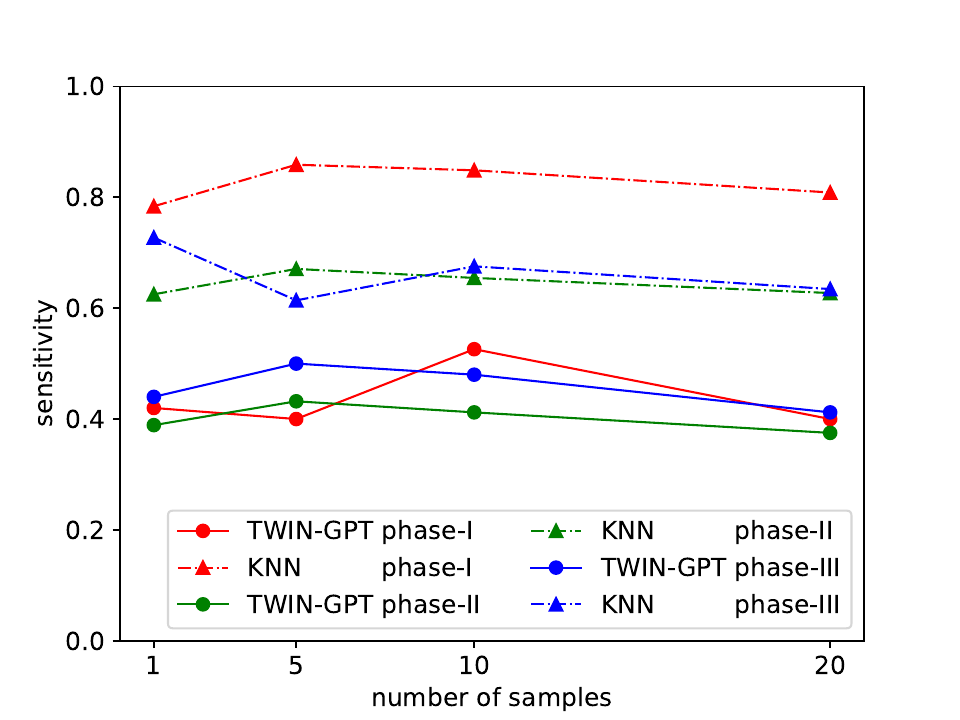}
  }
  \caption{On the \textit{TOP dataset}, presence disclosure sensitivity scores with a different number of samples known by the attacker. Lower sensitivity is better. }
  \label{fig:Presence_disclosure_TOP}
\end{figure}

\begin{figure}
  \centering
  \subfigure[Attribute disclosure: comprehensive perspective]{
  \includegraphics[width=0.48\textwidth]{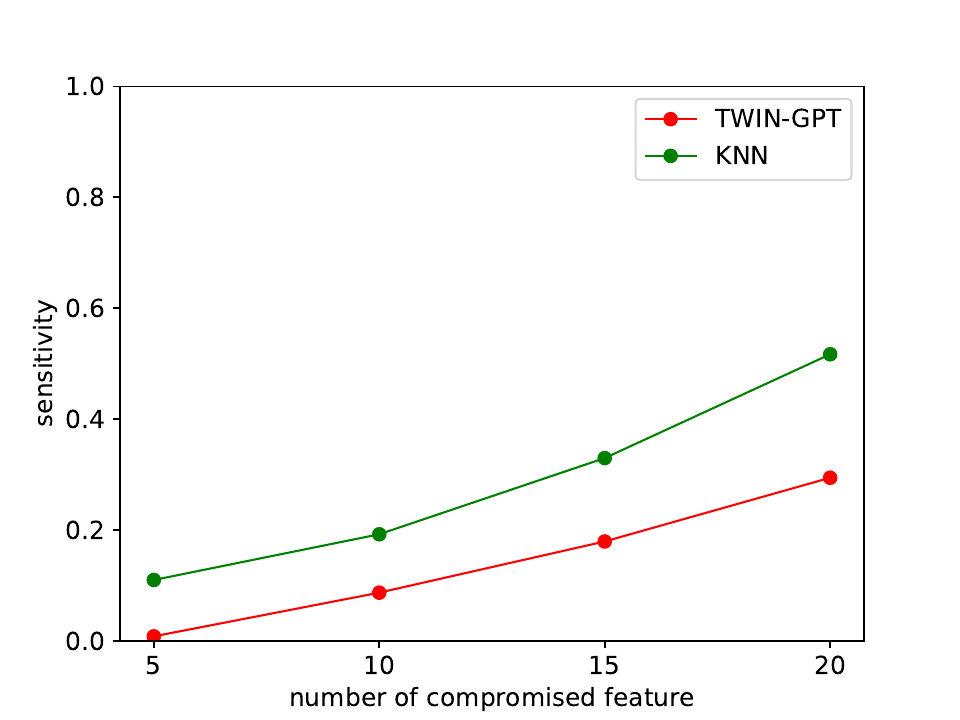}
  }
  \subfigure[Attribute disclosure: phase-by-phase analysis]{
  \includegraphics[width=0.48\textwidth]{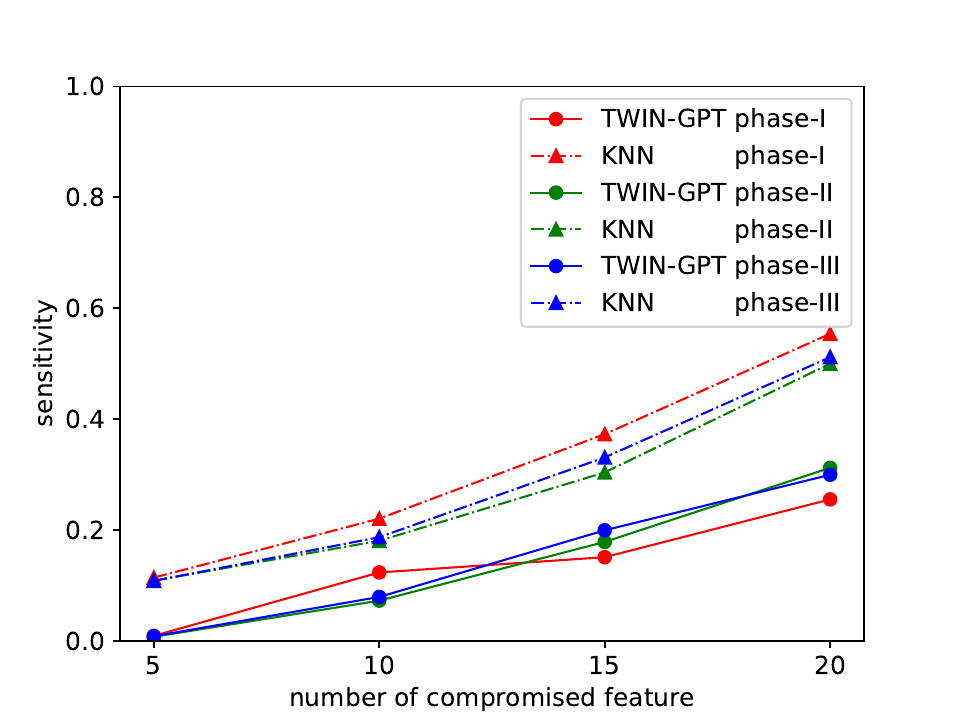}
  }
  \caption{On the \textit{TOP dataset}, Attribute disclosure sensitivity scores with a different number of samples known by the attacker. Lower sensitivity is better. }
  \label{fig:Attribute disclosure_TOP}
\end{figure}

\subsubsection{Presence Disclosure}
If an attacker finds that the synthetic data was trained by \texttt{TWIN-GPT} from the patient \textit{n}'s record $X_{n;1:T_{n}}$, we call it presence disclosure. We assume that there are total $m$\% of training data that had been known by the attackers and set $m$ as 1\%, 5\%, 10\%, and 20\%. After calculating the sensitivity through Eq. 7, we display the result in Fig. \ref{PandAttribute_oct}(a). As the number of known samples rises, the sensitivity of \texttt{TWIN-GPT} has remained stable at around 20\% (the attackers can recognize 20\% of the real patients' records from synthetic data they know). Correspondingly, the $k$-NN-based method reaches the maximum sensitivity of around 50\%. Additionally, we observe that in terms of presence disclosure, \texttt{TWIN-GPT}'s sensitivity is slightly higher than that of \texttt{TWIN-VAE} but significantly lower than the $k$-NN algorithm. This could be because \texttt{TWIN-GPT}, as a large language model-based method, might possess a greater capability to capture and replicate the subtle differences of the training data. Such high fidelity in data replication may inadvertently lead to synthetic data retaining too many identifiable features of the original training data, thus increasing the risk of presence disclosure. However, on the \textit{TOP dataset}, both the overall medical trial sensitivity for presence disclosure and the phase-by-phase medical trial sensitivity for presence disclosure are better than $k$-NN in Fig. \ref{fig:Presence_disclosure_TOP}.

\subsubsection{Attribute Disclosure}
Here, we assume that the attackers know parts of the training set and have access to $x$\% of features of the records. We set $x$ as 1, 5, 10, and 20 in this experiment. When an attacker can infer additional attributes of an individual based on the features of a subset of the data he knows, the attribute disclosure occurs. We use Eq. 8 to calculate mean sensitivity. The results are shown in Fig. \ref{PandAttribute_oct}(b). We can see that the mean sensitivity of \texttt{TWIN-GPT} is generally less than the $k$-NN-based method and Twin method, meaning better performance. We can also observe that the maximum score of $k$-NN-based method and \texttt{TWIN-GPT} are respectively around 0.3 and 0.25. On the \textit{TOP dataset}, as shown in Fig. \ref{fig:Attribute disclosure_TOP}, the mean sensitivity of \texttt{TWIN-GPT} is significantly lower than that of the $k$-NN algorithm. Across different numbers of features, \texttt{TWIN-GPT}'s mean sensitivity is 0.2 percentage points lower than that of $k$-NN. 

\subsubsection{Nearest neighbor adversarial accuracy risk}
NNAA is a measure of the degree to which a model overfits the original data, directly relating to privacy leakage risk. Here, we select 71 participants with 500 visit records as evaluation sets $S_E$. Correspondingly, we also choose 500 clinical records from real data as training sets $S_T$ and from synthetic data to form $S_S$, like the method required. \texttt{TWIN-GPT} achieves a score of 0.271. Generally, If the NNAA score is close to 0.5, we say that models overfit the original data. Therefore it suggests that \texttt{TWIN-GPT} is not excessively memorizing the original data and is instead learning more generalizable patterns.

\subsection{Model Explainability}
\begin{figure*}
  \centering
  \includegraphics[width=0.8\textwidth]{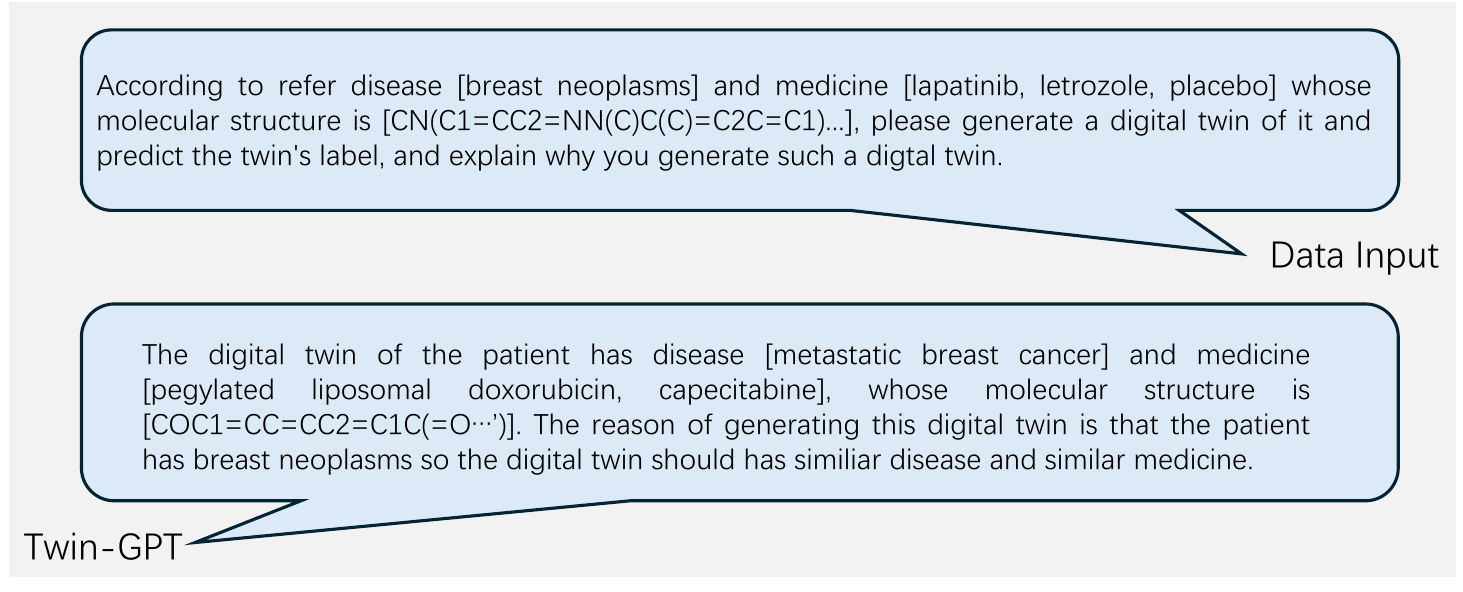}
  \caption{Explanation of \texttt{TWIN-GPT}. }
  \label{fig:explanation}
\end{figure*}

Unlike traditional models, our model functions as a ``world model'' endowed with a rich set of capabilities, enabling it to understand and interact with complex data landscapes. Therefore, our model possesses strong explainability. To this end, we conducted the explainability test of the twin generation results. As shown in Fig. \ref{fig:explanation}, \texttt{TWIN-GPT} can provide explanations for its results when generating twins from input data.

\subsection{Ablation Study}
We compare the performance of \texttt{TWIN-GPT} before prompt fine-tuning and after prompt fine-tuning. We refer to the non-fine-tuned version as \texttt{TWIN-GPT-origin}. The accuracy of \texttt{TWIN-GPT} and \texttt{TWIN-GPT-origin} in severe outcome prediction is 0.821 and 0.537, respectively. Moreover, the attribute disclosure of \texttt{TWIN-GPT} when $x$ set to 20 is less than 0.3, but in contrast, \texttt{TWIN-GPT-origin} reaches 0.530, meaning that attackers can easily infer additional attributes of an individual based on the features of a subset of the data he knows. In terms of the indicator of Personalized Generation quality and Counterfactual Generation quality, most of the participants' $r$ for TWIN-GPT are larger than 0.8, but the majority of participants' $r$ for TWIN-GPT-origin are less than 0.2.
\begin{table}[h]
\centering

\begin{tabular}{ccc}
\toprule 
\textbf{Metric} & \textbf{TWIN-GPT} & \textbf{TWIN-GPT-origin} \\ \midrule
Accuracy in Severe Outcome Prediction & 0.821 & 0.537 \\ \hline
Attribute Disclosure when $x = 20$ & 0.260 & 0.530 \\ \hline
Personalized Generation ($r$ value) &  0.812 & 0.121 \\ \hline
Counterfactual Generation ($r$ value) &0.785 & 0.310 \\ \bottomrule
\end{tabular}
\caption{Ablation study of \texttt{TWIN-GPT}  and \texttt{TWIN-GPT-origin}}

\label{tab:my_label}
\end{table}

\section{CONCLUSIONS}
In this paper, leveraging ChatGPT as the base model, we developed a specific large language model called \texttt{TWIN-GPT}, for generating personalized digital twins tailored for virtual clinical trial simulation. The approach proves advantageous for efficient and accurate prediction of clinical trial outcomes, particularly when confronted with limited EHR training data. The \texttt{TWIN-GPT} demonstrates exceptional performance, especially in terms of fidelity, utility, and privacy. The evaluation further validates that the \texttt{TWIN-GPT} can pose low privacy risks concerning presence disclosure, attribute disclosure, and nearest neighbor adversarial accuracy risks, effectively addressing privacy concerns inherent in traditional physical clinical trials.

\bibliographystyle{plain}
\bibliography{ref}

\end{document}